\documentclass[nonacm,sigconf,10pt]{acmart}
\usepackage{graphicx}
\usepackage{multirow}
\usepackage{subcaption}
\usepackage{xspace}
\usepackage{xcolor}
\usepackage{algorithm}
\usepackage{algorithmicx}
\usepackage[noend]{algpseudocode}
\usepackage{enumitem}
\usepackage{tabularx}
\usepackage{lscape}
\usepackage{makecell}
\usepackage{geometry}
\usepackage{array}

\AtBeginDocument{%
  }

\begin{document}
\settopmatter{printfolios=true}
\settopmatter{printacmref=false}
\setcopyright{none} 
\title{HyperCam: Low-Power Onboard Computer Vision for IoT Cameras}


\author{\texorpdfstring{
  Chae Young Lee, Pu (Luke) Yi, Maxwell Fite, Tejus Rao,\\
  Sara Achour, Zerina Kapetanovic \\
  \vspace{5pt}
  Stanford University, Stanford, USA}{Chae Young Lee, Pu (Luke) Yi, Maxwell Fite, Tejus Rao, Sara Achour, Zerina Kapetanovic, Stanford University}
}

\renewcommand{\shortauthors}{Lee et al.}


\begin{abstract}
We present HyperCam, an energy-efficient image classification pipeline that enables computer vision tasks onboard low-power IoT camera systems. HyperCam leverages hyperdimensional computing to perform training and inference efficiently on low-power microcontrollers. We implement a low-power wireless camera platform using off-the-shelf hardware and demonstrate that HyperCam can achieve an accuracy of $93.60\%$, $84.06\%$, $92.98\%$, and $72.79\%$ for MNIST, Fashion-MNIST, Face Detection, and Face Identification tasks, respectively, while significantly outperforming other classifiers in resource efficiency. \rev{Specifically, it delivers inference latency of 0.08-0.27s while using 42.91-63.00KB flash memory and 22.25KB RAM at peak.} Among other machine learning classifiers such as SVM, xgBoost, MicroNets, MobileNetV3, and MCUNetV3, HyperCam is the only classifier that achieves competitive accuracy while maintaining competitive memory footprint and inference latency that meets the resource requirements of low-power camera systems.
\end{abstract}

\newcommand{\prosehead}[1]{\noindent\textbf{\textit{#1}}}

\newcommand{\proseheadb}[1]{\noindent\textbf{#1}}

\newcommand{\ceil}[1]{\lceil #1 \rceil}

\definecolor{seafoamgreen}{HTML}{05c46b}
\definecolor{purple}{HTML}{8e44ad}
\definecolor{skyblue}{HTML}{3c40c6}

\newcommand{\hl}[1]{\textcolor{blue}{#1}}
\newcommand{\tool}[0]{HyperCam\xspace}

\newcommand{\blineDNN}[0]{\texttt{DNN}}
\newcommand{\blineEdgeDNN}[0]{\texttt{EdgeDNN}}
\newcommand{\btool}[0]{\texttt{\tool{}}}

\newcommand{\taskdetect}[0]{detect}
\newcommand{\taskid}[0]{identify}

\newcommand{\todo}[1]{\textcolor{red}{\textbf{TODO:} #1}}

\newcommand{\gray}[1]{{#1}}

\newcommand{\cmt}[1]{\textcolor{gray}{#1}}
\newcommand{\syn}[1]{\textbf{\textcolor{black}{#1}}}
\newcommand{\fun}[1]{\textit{\textcolor{black}{#1}}}

\newcommand{\reb}[1]{{#1}}
\newcommand{\rev}[1]{{#1}}


\maketitle

\section{Introduction}

Image sensors are now everywhere, found in smartphones, laptops, gaming consoles, and vehicles. Paired with advances in machine learning (ML), they enable object detection and classification for practical applications in areas such as healthcare, manufacturing, or transportation. However, ML models, especially deep neural networks (DNNs), require substantial computing power and memory, limiting the adoption of these techniques to Internet-of-Things (IoT). As a result, many IoT cameras offload images to the cloud or gateway servers with more computing resources \cite{neuricam, josephson2019wireless, naderiparizi2016wispcam, farmbeats}. However, this approach introduces overhead in latency, energy consumption, and data privacy. It also causes the system to depend on network communications, which can be costly and unreliable. Research shows that low-power camera systems frequently experience packet losses, leading to poor image quality at the base station \cite{naderiparizi2016wispcam, josephson2019wireless}.



\rev{An emerging alternative is onboard or embedded ML techniques, where computation occurs directly on sensor nodes. These systems can provide actionable insights in environments lacking reliable power and Internet connectivity. For example, in data-driven agriculture, embedded ML can analyze image data to assess crop yields or detect pests and plant diseases~\cite{farmbeats}. It can also support environmental and wildlife monitoring through camera traps or field survey robots~\cite{ahumada2020, Afzal2022underwater}. In these scenarios, there is often a lack of Internet connectivity or a low-bandwidth connection, where transmitting summaries of insights is more favorable than entire images. However, most existing embedded ML systems rely on DNNs, which are not well-suited to the tight constraints of microcontrollers (MCUs). Adapting these models involves techniques like quantization and pruning to meet memory and energy requirements. While these methods make deployment feasible, they frequently come at the cost of reduced accuracy~\cite{jacob2017quantization, banner2018nips}.}

\begin{figure}[t!]
    \centering
    \includegraphics[width=0.90\columnwidth]{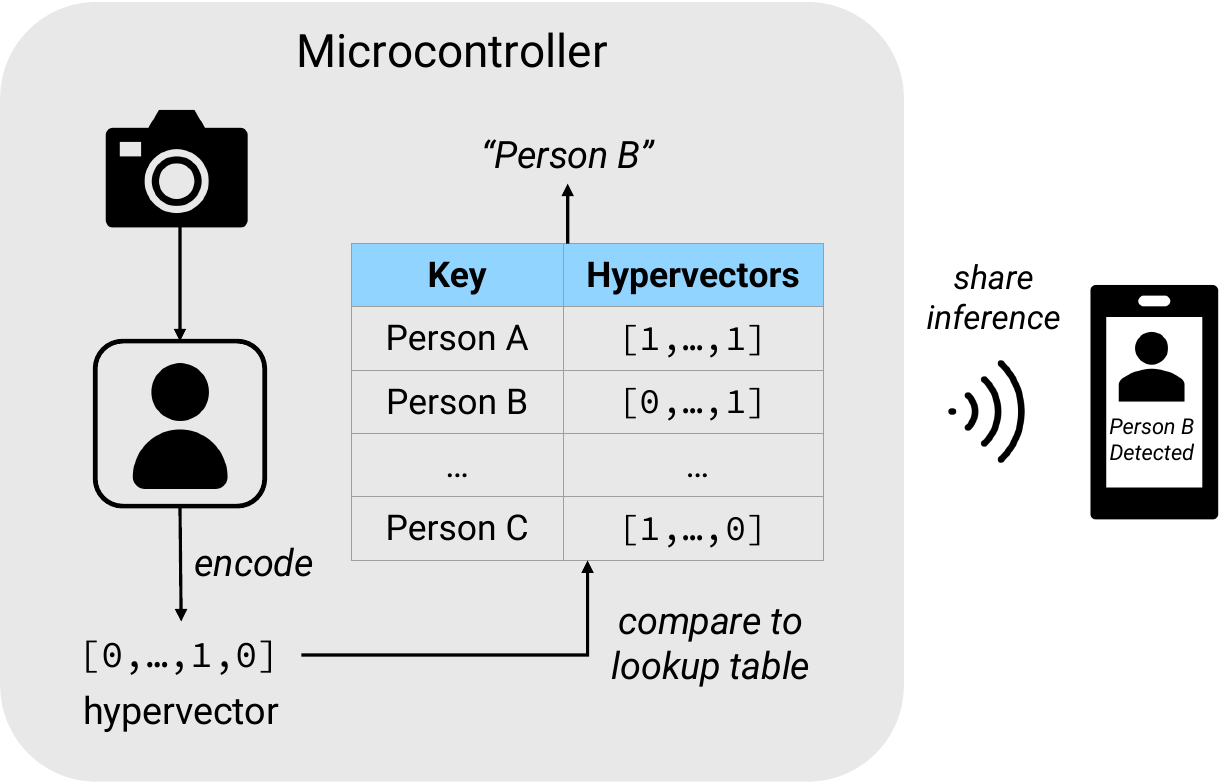}
    \caption{HDC for image classification. \normalfont{\tool{} uses an HD classifier to perform face detection and identification tasks onboard low-power wireless camera platforms.} }
    \label{fig:teaserfig}
    \vspace{-15pt}
\end{figure}

\begin{figure}
    \centering
    \includegraphics[width=0.5\columnwidth]{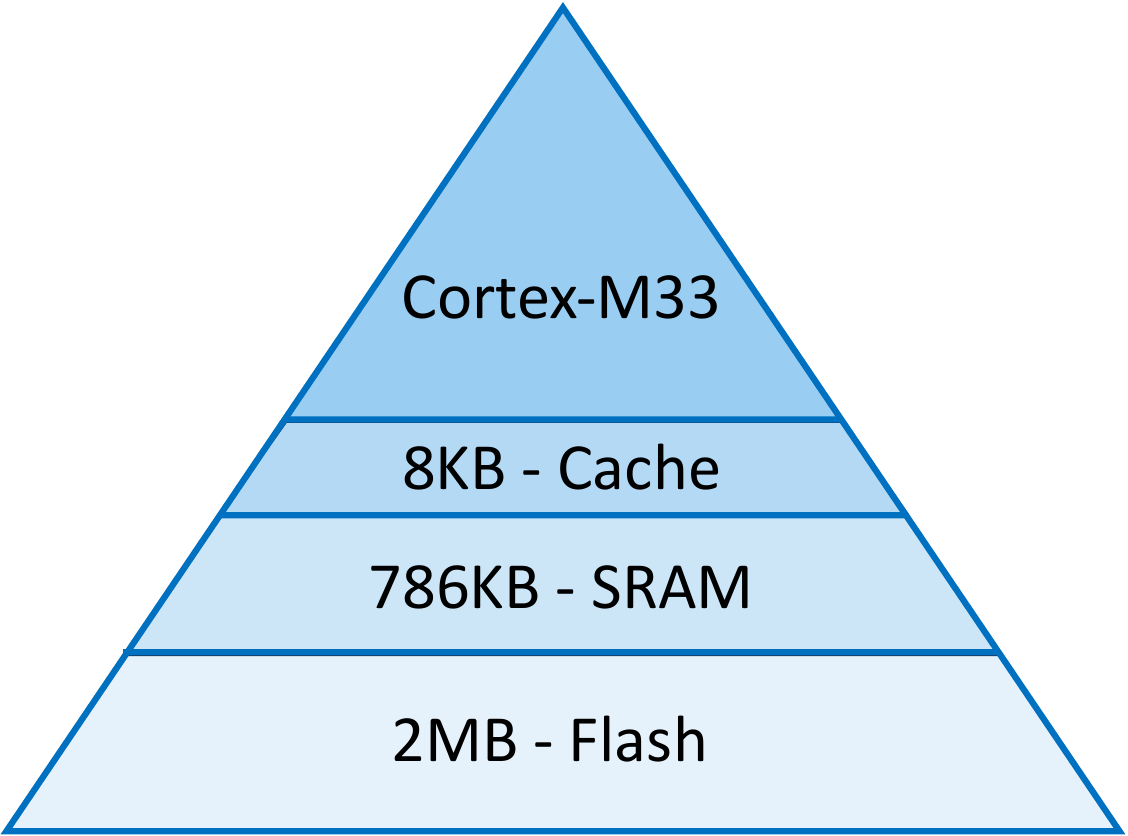}
    \caption{Memory layout of STM32U585AI.}
    \label{fig:memory-tier}
    \vspace{-15pt}
\end{figure}

This paper presents \tool, an innovative image processing pipeline designed for resource-constrained camera systems. As shown in Fig.~\ref{fig:teaserfig}, \tool{} processes images locally, classifies them in real-time using an onboard model, and transmits the results wirelessly to a nearby smartphone. At its core, \tool{} uses hyperdimensional computing (HDC), a computational paradigm based on structured data types and bitwise operations ~\cite{kleyko22}. Compared to DNNs, HDC is inherently hardware-friendly and energy-efficient \cite{imani2017exploring, langenegger2023memory}. However, most existing HDC works target time-series data, and image processing with HDC introduces unique challenges in memory and latency optimization. As shown in Fig.~\ref{fig:memory-tier}, MCUs typically have limited, flat memory hierarchies, and meeting these constraints requires careful model design. Additionally, optimizing latency is critical not only to meet real-time requirements but also to minimize the overall power consumption of the system. In image processing, the HD computation load increases proportionally to the image size. For example, a baseline HDC approach can take one minute to classify a $120 \times 160$ grayscale image.

\tool{} solves these challenges using novel and highly efficient HD encoding methods and aggressively optimizing performance in terms of memory and latency. Specifically, it features a lightweight encoder that dynamically maps images into HD space, eliminating the need for pre-stored mappings as other HD classifiers. The encoder also uses a sparse binary bundling based on Bloom Filter and Count Sketch, reducing the number of encoding operations by two orders of magnitude. \rev{Integrated into an ARM Cortex M-33 microprocessor, \tool{} is $21.08\%$ more accurate than MicroNet and $21.51\%$ more accurate than MobileNet-V3-Small in the 7-class face identification task. It is also 55-398 times faster and 12-33 times more lightweight than these baseline DNNs.} The following are key contributions made in this work.

\begin{itemize}[leftmargin=*]
    \item We introduce \tool, a novel HD image classifier that deploys highly efficient novel data encodings to perform inference on the sensor node. \tool is far more accurate, lightweight, and fast than previous HDC methods and DNN baselines.
    \item We develop a prototype of a low-power wireless camera platform to evaluate \tool.
    \item We show that \tool can perform binary and multiclass classifications in real time using captured image frames. \rev{\tool achieves an accuracy of $92.98\%$ and $72.79\%$ for face detection and identification, respectively, using less than 60 kilobytes of memory and achieving a latency of 0.27 seconds.} 
    \item We open source the \tool code to help promote reproducibility and advance onboard computing methods. 
\end{itemize}

\begin{figure*}[t!]
    \centering
    \includegraphics[width=0.9\textwidth]{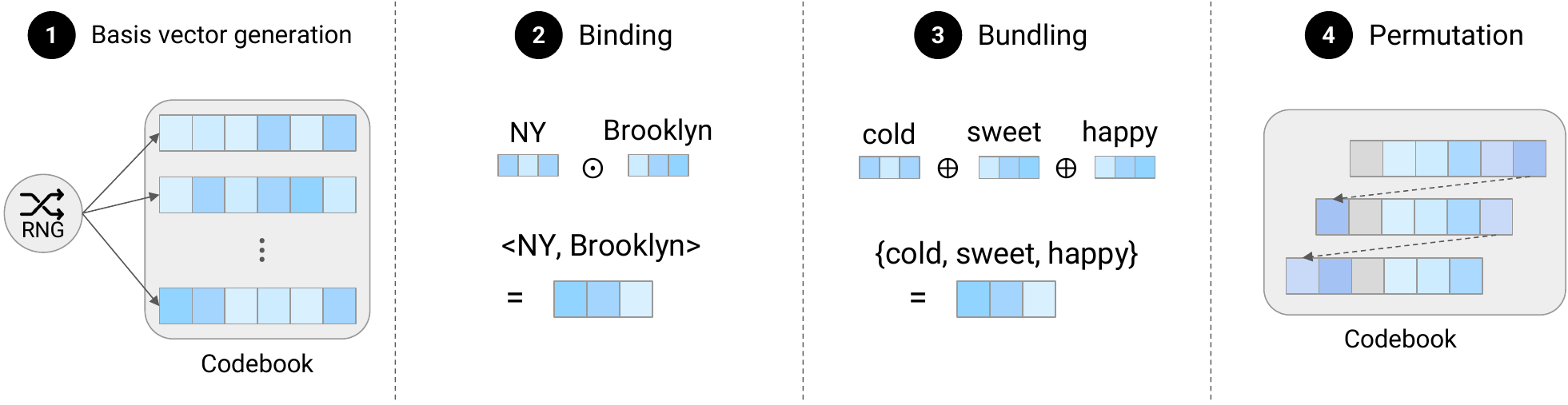}
    \caption{Key operations of BSC. \normalfont{\textbf{(1)} Basis vectors are generated for every letter. \textbf{(2)} Binding of data creates a record. \textbf{(3)} Bundling of words creates a set. \textbf{(4)} Permutation is applied to create hypervectors on-the-fly.}}
    \label{fig:bsc}
    \vspace{-7pt}
\end{figure*}

\section{Hyperdimensional Computing Background}

Hyperdimensional Computing (HDC), or Vector Symbolic Architectures (VSA), is a brain-inspired computing paradigm that represents information in a high-dimensional space. This framework encodes data as \textit{hypervectors}, vectors typically consisting of thousands of dimensions. Randomly generated hypervectors called the \textit{basis hypervectors} represent discrete data units such as symbols and numbers. Applying HDC operators such as binding, bundling, and permutation on these basis hypervectors constructs hypervector representations of more complex data structures (e.g., sequences, trees, and images). Information can be retrieved from hypervectors by computing the distances between hypervectors. HDC models vary widely in terms of their hypervector representations, operators, and distance metrics choices. \tool{} uses the Binary Spatter Code (BSC) approach~\cite{kanerva1997fully}, where each element of a hypervector is binary. Operations done on binary hypervectors are simple, energy-efficient, and thus, the best choice for resource-constrained hardware. In the following sections, along with Fig.~\ref{fig:bsc}, BSC-HDC operations are explained in more detail.

\subsection{Binary Spatter Code}

\subsubsection{Basis vector generation}
In BSC, each unique symbol is represented as a binary hypervector called the basis hypervector. Each vector element is a bit, randomly generated with a $p=0.5$ Bernoulli. The hyperdimensionality of these vectors ensures that randomly generated vectors are nearly orthogonal. In other words, any two basis hypervectors are far apart, usually with a Hamming distance of about 0.5. These basis hypervectors, also called \textit{codes}, are stored in a dictionary data type called the \textit{codebook}.

\subsubsection{Binding}

The binding operator $(\odot)$ combines basis hypervectors and creates a hypervector dissimilar to the input. In BSC, binding is implemented as an exclusive OR (XOR). Binding is used to construct larger data structures such as composite symbols, key-value pairs, and positional encoding from basis hypervectors. For example, binding the two hypervectors that represent the words \texttt{cold} and \texttt{water} results in a single hypervector for \texttt{cold water}. Similarly, binding the hypervectors for the key and the value creates a hypervector for the key-value pair.

\subsubsection{Bundling}\label{sec:bundling}

The bundling operator $(\oplus)$ aggregates multiple hypervectors and outputs a hypervector similar to the input. In BSC, bundling is executed through an element-wise majority vote. Given two or more input hypervectors, the number of zeros and ones are counted at each index, and the output hypervector chooses the majority value at that index. Bundling is used to create sets of symbols or data instances. For example, an image hypervector is created by bundling the hypervectors of its pixels. Similarly, a hypervector for a database record is created by bundling the hypervectors of its key-value pairs.

\subsubsection{Permutation}\label{sec:permutation}
The permutation operator $(p)$ is implemented as a circular shift, which creates a dissimilar hypervector far apart from the input. Because of this characteristic, permutation is used to create new basis hypervectors as an alternative to random generation. Additionally, permutation is used to encode the position data of sequences. For example, a bigram can be encoded by binding the permuted hypervectors of the characters. That is, binding occurs between the hypervector of the first character and the permuted hypervector of the second character. Similarly, the dimension of an image array can be represented using permutation. For 2-dimensional images, binding occurs between the hypervector of the row index and the permuted hypervector of the column index.

\subsubsection{Distance metric}

While binding, bundling, and permutation operators encode raw data into hypervectors, distance metrics are used to retrieve information from the hyperdimensional space. The lower the distance between two hypervectors, the more similar they are. In BSC, distance measurement is implemented using the Hamming distance, which counts the number of differing bits and normalizes the count by the length of the hypervector. The distance metric is often used to identify the class of the query hypervector. Other times, it decodes the hypervector to its raw data form (e.g., sequences, images). For example, the identity of the key in a hypervector for a key-value pair can be determined by computing the distance between the hypervector and all possible key hypervectors. 

\begin{figure*}[ht!]
    \centering
    \includegraphics[width=.8\textwidth]{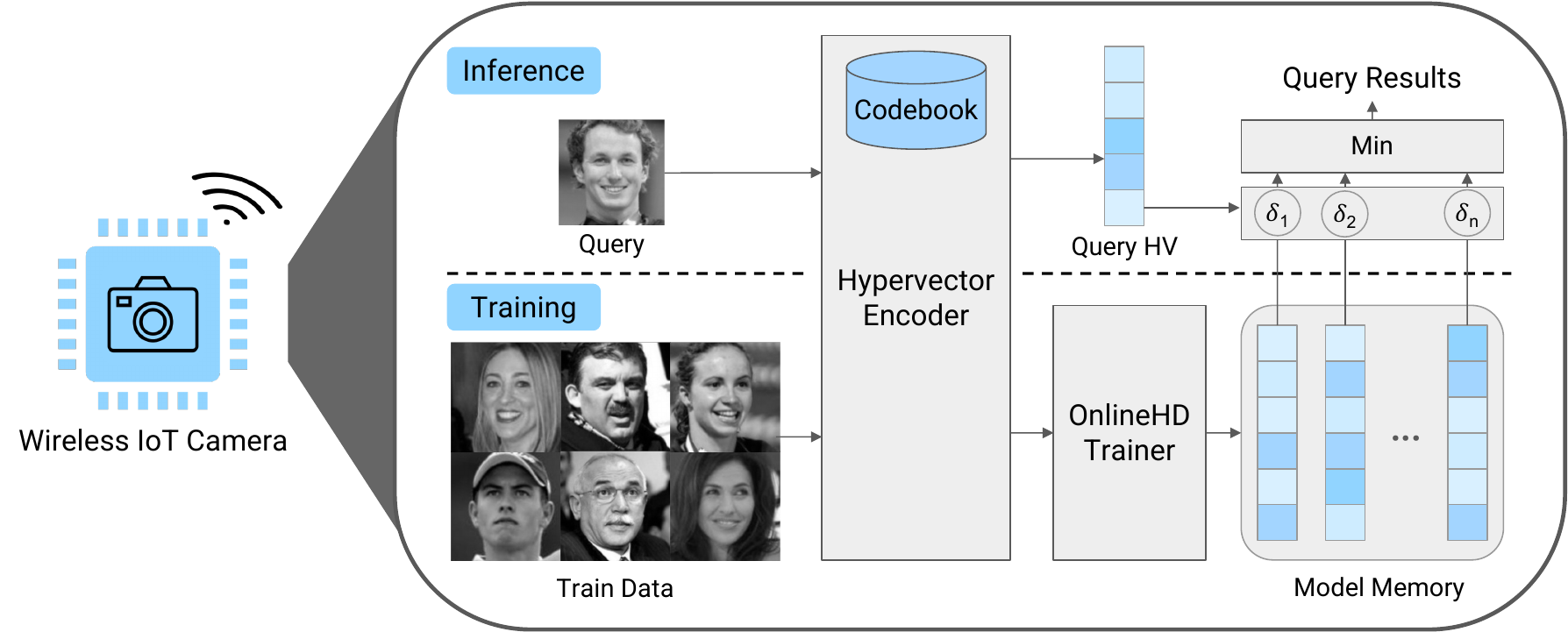}
    \caption{HyperCam overview. {\normalfont The \tool classifier runs onboard a low-power wireless camera platform and has three key components: image encoder, training algorithm, and inference algorithm. }}
    \label{fig:system_overview}
    \vspace{-7pt}
\end{figure*}

\section{\tool Design}

Using BSC-HDC operations, \tool processes computer vision tasks at the endpoint device (e.g., wireless IoT camera). When the camera captures an image, it is converted into a hypervector, and the classifier determines its class based on the Hamming distances. In addition, \tool{} can send the classification result to an IoT gateway via Bluetooth, allowing remote monitoring and interaction. This approach of transmitting the classification data, as opposed to entire raw images, significantly reduces communication overhead and mitigates the risk of transmission errors. The architecture of the \tool{} classifier, as shown in Fig.\ref{fig:system_overview}, has three major components: an image encoder, a training algorithm, and an inference algorithm.

\proseheadb{Image Encoder.} Both inference and training require that images be first encoded as hypervectors. This translation is done by performing an HD computation over basis hypervectors that capture pixel position and value information. The MCU stores codebooks that contain these basis hypervectors. A critical challenge in applying HD classifiers to image classification tasks is managing the performance and memory overhead associated with encoding image data as hypervectors. \tool{} deploys a novel image encoding algorithm (Section~\ref{sec:image-encoding}) that exploits the structure of HD computations to drastically reduce the memory usage and latency of the encoding procedure. This encoding algorithm uses a novel sparse bundling algorithm (Section~\ref{sec:sparse-bundling}) to accelerate the bundling of sets of elements.

\proseheadb{Training.} Training the \tool{}'s HD classifier occurs offline on a commodity computer. In this phase, training data are first encoded to hypervectors through the image encoder. Then, these encoded hypervectors are grouped based on their class labels. The class hypervector is construed by bundling together the hypervectors of all data instances that belong to that class. The table of class hypervectors is called the \textit{item memory}. While HD classifiers are typically trained using a one-shot algorithm, \tool{} uses the OnlineHD adaptive training algorithm ~\cite{onlinehd2021}. OnlineHD provides an effective few-pass learning approach where classifier hypervectors are refined based on the misclassifications observed on each training iteration. OnlineHD targets MAP-HDC, which works with real-valued vectors. \tool{} works with a modified version of OnlineHD that works with binary hypervectors. The adapted algorithm binarizes the real-valued classifier vectors after each training iteration and uses the binarized item memory to find misclassifications and update the real-valued model.

\proseheadb{Inference.} During execution, the MCU encodes each input frame to a hypervector. This hypervector, called the query hypervector, is compared to all the class hypervectors in the item memory using the Hamming distances. The class with the smallest distance to this query is the predicted category of the input. This inference computation is highly computationally efficient, involving only simple Hamming distance calculation.

\begin{table*}[t]
    \centering
    \begin{tabular}{|c|r|r|r|r|r|}
        \hline
         \multirow{2}{*}{\textbf{}} & \textbf{Naive} & \textbf{Rewrite 1} & \textbf{Rewrite 2} & \textbf{Rewrite 3} & \textbf{\tool{}}\\
         \hline
         \textbf{Codebook} & $536$ & $258$ & $258$ & $258$ & $258$ \\
         \textbf{Bind} & $38400$ & $38400$ & $19200$ & $19200$ & $19200$ \\
         \textbf{Bundle} & $19200$ & $19200$ & $19200$ & $19456$ & $256$ \\
         \textbf{SparseBundle} & $0$ & $0$ & $0$ & $0$ & $19200$ \\
         \hline
    \end{tabular}
    \caption{\rev{Comparison of encoding methods based on the size of the codebook and the number of bind, bundle, and sparse bundle operations. {\normalfont Each bundling operation involves 10000 bit-wise addition, whereas each sparse bundling operation involves 20.}}}
    \label{tab:encoding}
    \vspace{-15pt}
\end{table*}


\section{\tool{} Image Encoding}\label{sec:image-encoding}

\tool{}'s image encoding uses HD expression optimizations and a novel sparse bundling operator to reduce the encoding overheads dramatically. Section~\ref{sec:na\"{i}ve-encoding} presents the na\"{i}ve image encoding \tool{}'s encoding is based on, Section~\ref{sec:hypercam-encoding} presents the rewrites applied to reduce memory and computation requirements, and Section~\ref{sec:sparse-bundling} presents the novel sparse bundling method \tool{} uses to expedite image encoding. 

Table~\ref{tab:encoding} presents the computation and memory requirements of the unoptimized, na\"{i}ve encoding compared to the optimized encoding employed by \tool{}. The \tool{} encoding is obtained by applying four HD expression rewrites (\textit{Rewrite 1-4}) that progressively reduce the space and computational requirements of the encoder. \tool{} employs a novel sparse bundling algorithm that approximates HDC bundling while requiring 0.2\% of the operations. With these algorithmic encoding optimizations, \tool{}'s encoding algorithm uses 52\% less codebook memory, 50\% less binding operations, and 98\% less bundling operations. The 19200 sparse bundling operations in the final encoding are computationally equivalent to 38 normal HD bundling operations.

\subsection{Na\"{i}ve HD Image Encoding}\label{sec:na\"{i}ve-encoding}

This section describes a na\"{i}ve pixel-based HD image encoding algorithm for grayscale images. \tool{} works with a heavily optimized encoding derived from this na\"{i}ve encoding. In the following encoding formulations, $n$, $w$, and $h$ refer to hypervector size, image width, and height. \tool{} supports encoding 8-bit grayscale images, where each pixel value $Img[i,j]$ (or $Img[iw+j]$ in 1D format) is represented as an integer between 0 and 255.

\prosehead{Pixel Position Codebook.} Every pixel in a grayscale image has a unique position defined by its row and column indices. To encode this spatial information, we create two codebooks with randomly generated binary hypervectors. The pixel row codebook $R(i)$ maps the pixel rows $i \in 1\cdots{}h$, while the pixel column codebook $C(j)$ maps the pixel columns $j \in 1\cdots{}w$.

\prosehead{Pixel Value Codebook.} The pixel value codebook $V(x)$ maps an 8-bit grayscale value $x \in 0 \cdots{} 255$ to a hypervector. To ensure that similar pixel values have similar hypervectors, we use a level-based encoding technique~\cite{rachkovskiy2005sparse}. In this level-based codebook, $V(0)$ is instantiated to a zero vector of length $n$ representing a black pixel value. The basis hypervectors for values $1,\cdots, 255$ are constructed by sequentially setting random selections of zero-valued bits to one. 

\prosehead{Pixel Encoding.} To encode a single pixel located at $(i,j)$, we combine the spatial and intensity information. This is done by binding the hypervectors for the row $R(i)$, column $C(j)$, and pixel value $V(Img[iw+j])$:

\[
hv_{pix,i,j} = R(i) \odot C(j) \odot V(Img[iw + j])
\]

\prosehead{Image Encoding.} Once all pixel hypervectors are created, we aggregate them into a single hypervector that represents the entire image by bundling them:

\[
hv_{img} = \sum_{i=1}^h \sum_{j=1}^w R(i) \odot C(j) \odot V(Img[iw + j])
\]

\prosehead{Space and Time Complexity.} Given $n=10000$ bits, this encoding method needs to store $w + h + 256$ codebook hypervectors. It requires $wh$ pixel bundling operations and $2wh$ binding operations per image, each of which is a $n$-bit hypervector operation. For the $120 \times 160$ grayscale images used in this implementation, na\"{i}ve encoding would require $536$ codebook hypervectors totaling $670$ kilobytes of memory, 38400 binding operations, and 19200 bundling operations. Therefore, even for small images, this encoding algorithm scales poorly.

\subsection{\tool{} HD Image Encoding}\label{sec:hypercam-encoding}

Based on the na\"{i}ve encoding method presented in Section~\ref{sec:na\"{i}ve-encoding}, the properties of HD computations are exploited to rewrite the image encoding and optimize computation and memory usage. Sections~\ref{sec:r1}-\ref{sec:r4} present the HD expression rewrites applied to reduce the image encoder's memory footprint and computational requirements. The rewrites presented in~\ref{sec:r1}-\ref{sec:r2} preserve the computational properties of the HD encoding and therefore do not affect classification accuracy. The factoring and sparse bundling rewrites in~\ref{sec:r3}-\ref{sec:r4} are semantics-breaking and change the computational properties of the HD encoding, therefore affecting classification accuracy. \tool{}'s sparse bundling optimization uses a novel lossy filter-based sparse bundling operator, which is presented in Section~\ref{sec:sparse-bundling}.

\subsubsection{Rewrite 1: Permutation-based Codebooks}\label{sec:r1}

\tool{} uses the permutation operator (Section~\ref{sec:permutation}) to encode row and column indices, replacing the need for separate entries for each position in the row and column codebooks. Instead of storing $R(i)$ and $C(j)$ for all indices, the zero-index hypervectors $R(0)$ and $C(0)$ are repeatedly permuted to represent different rows and columns:

\[
hv_{img} = \sum_{i=1}^h \sum_{j=1}^w p^{i}[R(0)] \odot p^{\reb{j}}[C(0)] \odot V(Img[iw + j])
\]
Since the hypervectors in the pixel position codebook are generated independently and randomly, permuting one code effectively produces another independent code. In other words, any code and its permuted code have a high expected distance. This optimization reduces the number of codebook hypervectors from $w+h+256$ to $\ceil{\frac{w}{n}}+\ceil{\frac{h}{n}}+256$ hypervectors, cutting memory usage from 670 KB to 322 KB.

\subsubsection{Rewrite 2: Row and Column Index Coalescing}\label{sec:r2}

In the na"{i}ve encoding, row and column hypervectors are bound together to encode pixel positions. This binding step can be eliminated by introducing a new codebook $X$, which directly represents 1D pixel positions ($k=iw+j$):

\[
hv_{img} = \sum_{i=1}^h \sum_{j=1}^w X(iw + j) \odot V(Img[iw + j])
\]
This rewrite can be applied because the binding operator produces a hypervector that is dissimilar to its input hypervectors, and the input hypervectors are already independent. Replacing the 2D position encoding $(i,j)$ with a 1D pixel index $k$ preserves the behavior of the HD encoding while reducing unnecessary binding operations. The permutation from rewrite 1 can then be applied to reduce the codebook size from to $\ceil{\frac{wh}{n}}$ entries:

\[
hv_{img} = \sum_{i=1}^h \sum_{j=1}^w p^{i \cdot w + j}[X(0)] \odot V(Img[i w + j])
\]

\subsubsection{Rewrite 3: Value Hypervector Factoring}\label{sec:r3}

Critically, the number of bundling operations must be reduced to encode the image efficiently. The sparse bundling operator efficiently approximates bundling operations over permutations of a single hypervector. First, the value hypervector binding operations are factored from the bundling operation:

\[
hv_{img}  =   \sum_{z=0}^{255} V(z) \odot \left[\sum_{i,j \in Pix(z)} p^{iw + j}[X(0)] \right] 
\]
$Pix(z)$ returns all pixel positions $i,j$ where each pixel $Img[iw + j]$ has the value $z$. Note that the HD operation $\oplus$ is not associative since some information is lost during the quantization step in bundling. Thus, this rewrite changes the distance properties of the encoded hypervectors. Specifically, this rewrite loses information about the relative prevalence of different pixel values in the image. For example, if an image contains one gray pixel and many white pixels, the white and gray pixels would be equally important in this factored encoding. This information is re-introduced into the encoding using a weighted bundling: more prevalent pixel values are bundled multiple times.

\[
hv_{img}  =   \sum_{z=0}^{255}  |Pix(z)| \cdot V(z) \odot \left[\sum_{i,j \in Pix(z)} p^{iw + j}[X(0)] \right] 
\]
Here, values that occur more frequently in the image are heavily weighted in the encoding, recouping some information lost in the factored operation. This weighted bundling operation is equivalent to an HD bundling operation, where each hypervector is bundled multiple times:

\[
\sum_{z=0}^{255} |Pix(z)| \cdot [V(z) \cdots{}] 
= \sum_{z=0}^{255} \sum_{k}^{|Pix(z)|}  [V(z) 
 \cdots{}] 
\]
Therefore, the weighted bundling operation can easily be fused with a normal bundling operation by scaling the binary hypervector during the sum-threshold computation.

\subsubsection{Rewrite 4: Sparse Bundling}\label{sec:r4}

After applying the factoring rewrite, each pixel bundling sub-computation (blue text) can then be efficiently approximated using a novel sparse bundling operator introduced in Section~\ref{sec:sparse-bundling}:

\[
hv_{img}  =   \sum_{z=0}^{255} \left|Pix(z)\right| \cdot V(z) \odot \hl{\left[\sum_{i,j \in Pix(z)} p^{iw + j}[X(0)] \right]} 
\]
The sparse bundling operation approximates the standard HD bundling and replaces bundling operations over permuted hypervectors. It is designed to preserve the property of bundling that similar sets of pixels are embedded into hypervectors that are close to each other. It also processes a set of elements (in this case, pixel positions) and returns a hypervector that approximates the distance properties of the standard bundled set of elements:

\begin{equation}\label{eqn:encoding}
hv_{img}  =   \sum_{z=0}^{255} |Pix(z)| \cdot V(z) \odot SparseBundle(Pix(z))
\end{equation}
The sparse bundling operator works with a density parameter $d$, where $d \ll n$, and performs $O(d)$ operations to bundle two hypervectors. Using a sparse bundling operation reduces the number of operations required to bundle each vector from $O(n)$ to $O(d)$. \tool{} uses $d=20$, thus reducing the number of operations per bundling operator from $10000$ to $20$ operations. \rev{$d$ is determined experimentally, where for $d<20$, \tool experiences sharp drop of accuracy.} Once sparse bundling is applied to construct each pixel set hypervector, \tool{} applies 256 weighted HD bundling operations to construct the final hypervector representation of the image.

\subsection{Sparse Bundling}\label{sec:sparse-bundling}

\begin{algorithm}
    \caption{Sparse Bundling Algorithm}
    \label{alg:sparse}
    \begin{algorithmic}[1]
    \State \syn{bool} $bloom$ = $false$; \cmt{//use a bloom filter or count-sketch}
    \State \syn{uint} $n$ = 10000; \cmt{// hypervector size}
    \State \syn{uint} $d$ = 20; \cmt{// density - the number of hashes per bundle}
    \State \cmt{// random set of size $d$ from values $\{0,1,\cdots,n-1\}$}
    \State \syn{uint8}[d] $indices \gets$ rand(0,n,size=d,replace=False); 
    \State \cmt{// random vector with values $\{-1,1\}$}
    \State \syn{int8}[d] $CS \gets$ rand([-1,1],size=d);
    \Function{SparseBundleElem}{hv,$s$}
        \For{$j$ in $0...d-1$}
            \State k = (indices[j]+$s$) \% n
            \If{bloom}
            \State hv[k] = 1
            \Else
            \State hv[k] = hv[k] + CS[j]
            \EndIf
        \EndFor{}
    \EndFunction{}
    \Function{NewSparseHV}{}
    \State \syn{int8}[n] hv = zeroes(n);
    \State \syn{return} hv;
    \EndFunction{}
    \Function{FinalizeHV}{hv}
    \If{$\neg$ bloom}
    \For{$i$ in $0..n$}
        \State hv[i] = 1 ? hv[i] >= 0 : 0
    \EndFor
    \EndIf{}
    \EndFunction{}
    \Procedure{SparseBundle}{S}
    \State hv = \fun{NewSparseHV}()
    \For{$s \in S$}
        \State \fun{SparseBundleElem}(hv,s)
    \EndFor
    \State \fun{FinalizeHV}(hv);
    \State return hv;
    \EndProcedure{}
    \end{algorithmic}
\end{algorithm}

\tool{} deploys a novel sparse bundling algorithm that uses Bloom Filter~\cite{bloom1970space} and Count Sketch~\cite{charikar2002finding} to approximately bundle large numbers of hypervectors together at low latency. Bloom Filters and Count Sketches are probabilistic data structures adept at representing sets of elements. Both data structures work with numeric vectors and are updated by randomly sampling and updating bits.
They can be viewed as a sub-class of HDC/VSA as they also compute in superposition~\cite{kleyko2020autoscaling,kleyko2022vector,clarkson2023capacity}.

Given a set of integers $s \in S$, the sparse bundling algorithm returns a binary hypervector that approximates bundling together the basis hypervectors that represent each element:

\[
    SparseBundle(S) \approx \sum_{s \in S} p^{s}[hv]
\]
The sparse bundling operation approximates an HD bundling operation over permutations ($p^s$) of some hypervector $hv$. The algorithm is parametrized with a hypervector size $n$ and density parameter $d$ and offers both Bloom Filter and Count Sketch backends. The Bloom Filter backend is more computationally efficient but less accurate than the Count Sketch backend. Each sparse bundling operation requires $d$ operations, significantly reducing the number of operations per bundling task when $d << n$.

\subsubsection{Algorithm Description}

This section describes Alg.~\ref{alg:sparse}. Given a set of integer values $S$ to bundle, the $SparseBundle$ operator instantiates a new sum hypervector (Line 23), uses the sparse bundling operator to add each value to the sum hypervector (Lines 24-25), and then finalizes the sum hypervector (Line 26) to obtain a binary hypervector that approximates the bundled result. The $SparseBundleElem$ routine updates the sum hypervector to bundle an integer element. 

\prosehead{Instantiation and Finalization.} The sum hypervector is instantiated to an $n$-dimensional signed integer vector comprised of zeroes. On finalization, each element is binarized by thresholding the value with zero to produce an n-dimensional binary vector. Finalization is only required for sum hypervectors in the Count Sketch backend; the Bloom Filter backend directly produces binary vectors.

\prosehead{Bundling}[Lines 8-14] The algorithm updates the sum hypervector $hv$ to include the integer $s$ by computing $d$ random indices from the integer value and then updating the values in these indices. For the Bloom Filter backend, each update sets the hypervector value to one. For the Count Sketch backend, the bundle hypervector value is randomly incremented or decremented. \tool{} precomputes the random indices, along with the random increment and decrement operations, and stores the values in the Count Sketch (CS) array.

\section{\tool{} Implementation}

This section describes the implementation of \tool{}. Section~\ref{sec:imp-encoding} presents the implementation of the image encoding algorithm and further engineering efforts to port the model onboard. Section~\ref{sec:data} describes the collection of the image dataset used to evaluate \tool{}. Lastly, Section~\ref{sec:hardware} describes \tool{}'s low-power hardware platform.


\subsection{Image Encoding Algorithm}\label{sec:imp-encoding}

\begin{algorithm}
\caption{Image Encoding with Sparse Bundling}
    \label{alg:sparse:implementation}
    \begin{algorithmic}
    \Function{encodeImage}{Img: image}
        \State \syn{int}[256][n] hvs;
        \State \syn{uint}[256] cnts;
        \State \syn{uint8}[n] imgHV;
        \For{$v$ in $0..256$}
        \State sumHVs[v] = NewSparseHV()
        \State cnts[v] = 0
        \EndFor
        \For{$k$ in $0..w \cdot h$}
            \State v = Img[k]
            \State SparseBundleElem(sumHVs[v], k)
            \State cnts[v] += 1
        \EndFor
        \For{$v$ in $0..255$}
            \For{$i$ in $0 \cdots{} n-1$}
                \State imgHV[i] += cnts[v] $\cdot$ (sumHVs[v][i] $xor$ getValueCB(v,i))
            \EndFor
        \EndFor
        \For{$i$ in $0 \cdots{} n-1$}
            \State imgHV[i] = 1 ? imgHV[i] > |wh|/2 : 0;
        \EndFor{}
        \State \syn{return} imgHV;
    \EndFunction{}
    \end{algorithmic}
\end{algorithm}

Alg. \ref{alg:sparse:implementation} presents the algorithm for computing the optimized image encoding presented in Equation~\ref{eqn:encoding}.

First, a single pass is taken over the input image, during which the position hypervectors are bundled using either a Count-Sketch-based or Bloom-Filter-based bundling operation. The binary hypervectors produced by the sparse bundling operation are then bound with the value hypervectors and bundled to form the final image hypervector, as described in Equation~\ref{eqn:encoding}. Each bundled vector is bound with the corresponding value hypervector from the codebook, resulting in $256$ hypervectors. These hypervectors are then bundled together, using weights equivalent to the number of pixels in each bin. Moreover, since each hypervector contains only $d$ non-zero elements, the vector summation in the binning process computes $d$ integer elements instead of $n$.

To further reduce image encoding time and eliminate value codebook, value hypervectors are generated on-the-fly instead of pre-storing them. As described in Section~\ref{sec:na\"{i}ve-encoding}, the value codebook uses level-based encoding, where random selections of bits in $V(0)$ are flipped. For a value $v$, $V(v)$ is generated by flipping $v \cdot \left\lfloor n/256\right\rfloor$ bits. The order in which bits are flipped must be the same at every generation for encoding consistency. Thus, this ordering of bit flips (which are indices of the $n$-length array) is stored in the microcontroller. Additionally, the generation of value hypervectors does not create overhead in computation because it integrates into the binding operation, which already iterates over the vector.



\prosehead{Complexity.} These implementation-level optimizations reduce the codebook size from $256+2$ to $2$ hypervectors. The above algorithm requires $256$ bundling operations, $256$ binding operations, and $19200$ sparse bundling operations. Each sparse bundling operation uses approximately 500 times fewer operations than standard HDC bundling.

\begin{figure}[t!]
    \centering
    \includegraphics[width=.7\columnwidth]{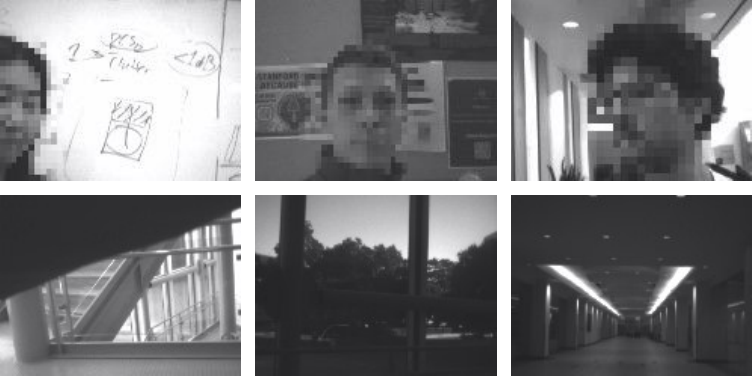}
    \caption{Sample images in the collected dataset.}
    \label{fig:dataset-collection}
    \vspace{-10pt}
\end{figure}



\subsection{Data Collection}\label{sec:data}
\rev{In IoT deployments, collecting data from real-world scenarios is crucial for aligning with actual conditions. IoT systems are sensitive to environmental changes, sensor noise, and operating conditions (e.g., lighting and object placement), affecting performance. We demonstrate an IoT deployment scenario by collecting a custom classification dataset and training a model.}
Specifically, we took 4,215 160x120 grayscale images using the Himax HM01B0 camera mounted on the Arducam HM01B0 Monochrome SPI Module. An assortment of backgrounds and people was imaged to diversify the input dataset. Images of people were taken such that their faces were captured at different angles and positions. Approximately 500 images per person were collected from various background scenes such as a hallway, office space, whiteboard, etc. Objects and backgrounds not involving people were also collected as negative samples for face detection. \reb{There are 4,215 images across seven person classes and one non-person class} as shown in Fig.~\ref{fig:dataset-collection}.

\subsection{Hardware}\label{sec:hardware}
A low-power camera hardware platform was designed to evaluate the performance of \tool on resource-constrained hardware. An evaluation board for the STM32UF855AI microcontroller (MCU) was used as the central computing device~\cite{B-U585I-IOT02A, STM32U585AI}. The MCU has an Arm Cortex-M33 processor, 2MB of flash memory, and 736KB of SRAM. The evaluation board contains several sensors, extra memory, and redundant peripheral interfaces for the evaluation of \tool. Thus, all non-critical components operating on the same power supply rails as the MCU were removed from the board to reduce power consumption. The Himax HM01B0 image sensor in QQVGA mode is used to capture 160x120 resolution grayscale images~\cite{himax}. A custom printed circuit board (PCB) is implemented to interface the MCU with the image sensor, which connects the 2.8V supply from the evaluation board to the camera. Moreover, it connects I2C and 8-bit parallel QQVGA communications between the MCU and image sensor. A 24MHz crystal oscillator drives the image sensor's internal clock. Lastly, the MCU's Digital Camera Interface (DCMI) and Direct Memory Access (DMA) peripherals are used to transfer image data from the camera into the MCU's memory. 

\begin{figure}[t]
    \centering
    \includegraphics[width=.7\columnwidth]{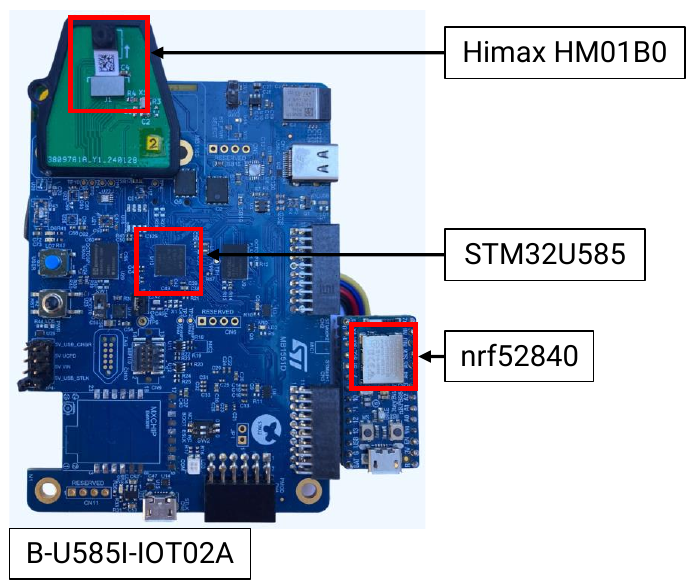}
    \caption{Low-power wireless camera platform.}
    \label{fig:hardware}
    \vspace{-10pt}
\end{figure}

\begin{table*}[!ht]
    \centering
    \small 
    \setlength{\tabcolsep}{3pt} 
    \renewcommand{\arraystretch}{1.5}
    \caption{\reb{Comparison of Image Classifiers. \normalfont{\tool is compared with different image classifiers in terms of accuracy (\%), flash memory usage (KB), peak RAM memory usage (KB), and latency (s) on 4 benchmark tasks.}}}
    \label{tab:classifier}
    \scalebox{0.8}{ 
    \begin{tabular}{|c|l|rrrr|rrrr|rrrr|rrrr|}
    \hline
        \multirow{2}{*}{\textbf{Type}} & ~ & \multicolumn{4}{c|}{\reb{\textbf{MNIST}}} & \multicolumn{4}{c|}{\reb{\textbf{Fashion MNIST}}} & \multicolumn{4}{c|}{\textbf{Face Detection}} & \multicolumn{4}{c|}{\textbf{Face Identification}} \\ \cline{3-18}
        ~ & ~ & \reb{\textbf{Acc}} & \reb{\textbf{Flash}} & \textbf{\reb{RAM}} & \reb{\textbf{Latency}} & \textbf{Acc} & \textbf{Flash} & \textbf{\reb{RAM}} & \textbf{Latency} & \textbf{Acc} & \textbf{Flash} & \textbf{\reb{RAM}} & \textbf{Latency} & \textbf{Acc} & \textbf{Flash} & \textbf{\reb{RAM}} & \textbf{Latency} \\ \hline
        \multirow{5}{*}{\textbf{HDC}} & \reb{VanillaHDC} & 80.03 & - & - & - & 69.39 & - & - & - & 72.54 & - & - & - & 40.60 & - & - & - \\
        ~ & \reb{OnlineHD} & 91.34 & - & - & - & 81.83 & - & - & - & 84.62 & - & - & - & 84.62 & - & - & - \\
        ~ & \reb{Rewrite 2} & 94.60 & 365.02 & 22.09 & 0.21 & 84.99 & 365.02 & 22.09 & 0.21 & 94.09 & 356.50 & 22.09 & 11.56 & 78.63 & 362.60 & 22.09 & 11.56 \\
        ~ & HyperCam$^*$ & 93.60 & 63.00 & 22.25 & 0.26 & 84.06 & 63.00 & 22.25 & 0.26 & 92.98 & 53.83 & 22.25 & 0.27 & 72.79 & 59.52 & 22.25 & 0.27 \\
        ~ & HyperCam$^{**}$ & 90.36 & 52.62 & 22.25 & 0.08 & 83.10 & 52.62 & 22.25 & 0.08 & 92.73 & 42.91 & 22.25 & 0.12 & 61.40 & 49.00 & 22.25 & 0.12 \\ \hline
        \multirow{2}{*}{\shortstack{\textbf{Lightweight} \\ \textbf{ML}}} & \reb{SVM} & 78.24 & - & - & - & 72.06 & - & - & - & 86.45 & - & - & - & 27.07 & - & - & - \\ 
        ~ & \reb{xgBoost} & 76.86 & 365.55 & 77.09 & 0.01 & 71.76 & 352.76 & 77.09 & 0.01 & 94.46 & 134.92 & 77.09 & 0.01 & 38.88 & 193.24 & 77.09 & 0.01 \\ \hline
        \multirow{4}{*}{\shortstack{\textbf{Neural}\\ \textbf{Networks}}} & MicroNets & 97.82 & 582.16 & 302.87 & 1.05 & 86.84 & 582.16 & 302.87 & 1.05 & 92.86 & 581.12 & 502.87 & 6.64 & 51.71 & 581.76 & 502.87 & 6.64 \\ 
        ~ & MobileNet V3 & 98.69 & 1640.00 & 302.87 & 3.29 & 86.48 & 1640.00 & 302.87 & 3.29 & 88.18 & 1640.00 & 502.87 & 18.53 & 51.28 & 1640.00 & 502.87 & 18.55 \\ 
        ~ & \reb{MCUNet V3$^{*}$} & 99.34 & 1190.00 & 302.91 & 6.70 & 93.3 & 1190.00 & 302.91 & 6.70 & 99.88 & 1190.00 & 302.91 & 6.70 & 99.15 & 1190.00 & 302.91 & 6.70 \\ 
        ~ & \reb{MCUNet V3$^{**}$} & 98.97 & 1340.00 & 502.91 & 46.71 & 94.20 & 1340.00 & 302.91 & 46.71 & 99.88 & 1340.00 & 502.91 & 46.71 & 99.01 & 1340.00 & 502.91 & 46.71 \\ \hline
    \end{tabular}}
\end{table*}

A nRF52840 BLE module is integrated into the camera hardware and wirelessly transmits data packets to a nearby base station~\cite{nRF52840}. A 3.7V 4400mAh Lithium Ion battery powers the entire hardware platform. Two linear regulators on the evaluation board provide 3.3V and 2.8V to the MCU and camera, respectively. A 3.3V linear regulator also supplies the BLE module. Some power losses were incurred in these linear regulators during the active state, which can be reduced by custom power management design.


The camera platform is designed so that components are set to standby in their minimum-sleep modes and are activated by an event trigger (e.g., motion detection or manual button press). When triggered, the MCU wakes the camera module to capture and store an image, performs image classification, and transmits the classification outcome to a smartphone application. After completing each task, the camera platform returns to sleep mode. Figure~\ref{fig:hardware} shows the prototype implementation of the hardware platform.

\section{Evaluation}
\tool{} is compared with baseline \reb{machine learning classifiers} in an identical embedded hardware environment.

\subsection{Experimental Setup}

\subsubsection{Classifier Tasks}

\reb{\tool is evaluated on four image classification tasks: MNIST, Fashion MNIST, Face Detection, and Face Identification. MNIST and Fashion MNIST are widely used benchmark datasets for evaluating machine learning classifiers, each containing 60,000 28x28 grayscale images~\cite{lecun1998mnist, xiao2017fashion}. Additionally, the Face Detection and Identification tasks utilize the collected dataset described in Section~\ref{sec:data}, which consists of 8 classes: 1 non-person class (objects and places) and 7-person classes. The Face Detection task is a binary classification distinguishing between the non-person class and the person class, while the Face Identification task classifies 7 person classes. All class sizes were balanced. Here, the benchmark datasets (MNIST and Fashion MNIST) are used to compare general approaches. In contrast, the custom tasks demonstrate an IoT deployment scenario where models handle lower quality and smaller datasets.}

\vspace{-10pt}
\subsubsection{Classifiers}

\reb{Several machine learning algorithms are selected as a baseline to compare against \tool{}. Both \tool{} and the baseline models are trained offline on a standard laptop, where their test accuracies are assessed. The trained models are then exported as C header files and loaded onto STM32U585AI for performance evaluation. All models use integer representations to fit the hardware and ensure compatibility with other MCU families. Except for the HDC models, which are inherently integer models, all other ML models were trained using floating-point numbers and then quantized post-training to integer values.}



\noindent\reb{\textbf{HDC.} Three baseline HDC models are assessed against two versions of \tool{}. VanillaHDC is the most basic form of an HD classifier explained in Section~\ref{sec:na\"{i}ve-encoding}. OnlineHD uses the OnlineHD~\cite{onlinehd2021} training method on top of VanillaHDC. Rewrite2 uses the encoding method described in Section~\ref{sec:r2} and uses the OnlineHD training method. On the other hand, \tool{} is assessed in two versions, HyperCam$^*$ with Count Sketch backend and HyperCam$^{**}$ with Bloom Filter backend. Both \tool{} models use the OnlineHD training method as well. Here, all hypervectors have a length of $n=10,000$.}

\noindent\reb{\textbf{Lightweight ML.} SVM and XGBoost are chosen to represent lightweight ML models. They are trained using Python's sklearn and xgboost libraries and are ported to a C header file using the micromlgen library.}

\noindent\reb{\textbf{Neural Networks.} MicroNets, MobileNetV3, and two sizes of MCUNetV3 are chosen for this category. MCUNet V3* (mcunet-in1) is the smallest, and the MCUNet V3** (mcunet-in3) is the largest one that fits the MCU. After training, they are trained from pre-trained weights and are quantized to 8-bit integer numbers. Once converted to C header files, the TensorFlow Lite Micro and the CMSIS-NN libraries are used to run them on the ARM Cortex M-33 environment.}

\vspace{-10pt}
\subsubsection{Evaluation Metrics.}These metrics were used:

\noindent\textbf{Accuracy.} Data is split in an 8:2 ratio between the training and testing datasets. The model is trained with the training dataset, and accuracy is measured using the test dataset.

\noindent\reb{\textbf{Flash Memory.} The flash memory footprint of the model is measured in kilobytes. For ML models, this includes the model weights, parameters, and the library code required for execution. For HDC models, this includes the model's codebook and the item memory.}

\noindent\reb{\textbf{RAM.} The peak RAM footprint of the model is measured in kilobytes. This includes the model activations, input, output tensors, and library code for ML models. For HDC models, this includes hypervectors allocated for encoding. When encoding is done, HDC uses only one hypervector to represent a data instance for inference.}

\noindent\textbf{Latency.} The latency of the classifier is the time it takes to process one frame of image. This involves the time it takes to encode an image and predict its class using the item memory. All latency is measured on STM32U585AI and is in seconds.

\subsection{Classifier Evaluation}\label{sec:evaluation}

Table \ref{tab:classifier} compares \tool{}'s HD classifier to the baseline classifiers in terms of accuracy, flash memory size, peak RAM size, and latency during one pass of inference.

\begin{figure}[b!]
    \centering
    \includegraphics[width=.8\columnwidth]{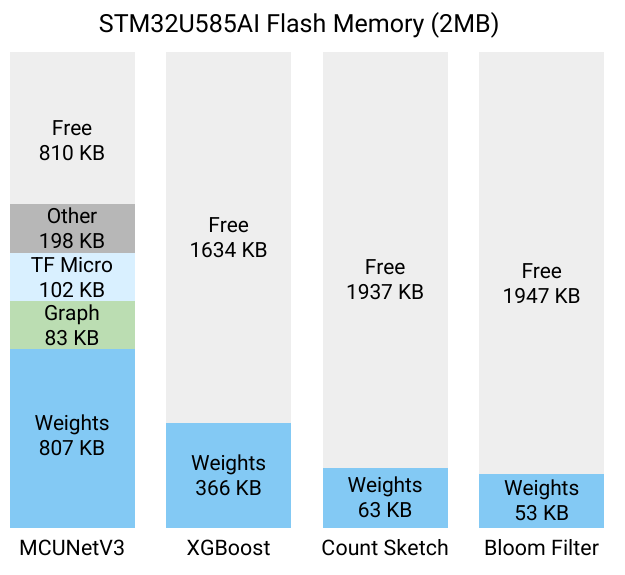}
    \caption{Breakdown of flash memory.}
    \label{fig:memory_breakdown}
    \vspace{-15pt}
\end{figure}

Among HD classifiers, VanillaHDC and {OnlineHD}, while demonstrating reasonable accuracy (80.03\% and 91.34\% on MNIST, respectively), are not suitable for deployment on resource-constrained devices due to their large flash memory footprint. The {Rewrite2} encoding method, proposed as part of \tool{}, significantly reduces the flash memory consumption to 365.02 KB for the largest task while maintaining a competitive accuracy of 94.60\% on MNIST and 84.06\% on Fashion MNIST. The final version of \tool{} further improves this by achieving the lowest flash memory footprint of all ML classifiers: 63.00 KB (HyperCam$^*$) and 52.62 KB (HyperCam$^{**}$) for the largest task. For a more competitive memory footprint and latency, \tool{} sacrifices accuracy from {Rewrite2} but only with a small margin (1.00\% reduction in MNIST and 0.93\% in Fashion MNIST). Furthermore, HyperCam$^{**}$ achieves the lowest latency across all HDC and neural network classifiers: 0.08 seconds on MNIST and 0.12 seconds on Face Detection and Identification.

\reb{When compared to lightweight machine learning models like {SVM} and {xgBoost}, \tool{} demonstrates superior performance in both accuracy and memory efficiency. For example, both versions of \tool{} achieve higher accuracy than {SVM} across all classification tasks, while {xgBoost} only outperforms \tool{} in the Face Detection task by a small margin of 1.48\%. In the Face Identification task, {SVM} and {xgBoost} experience a significant drop in accuracy (27.07\% and 38.88\%, respectively). By contrast, all HD classifiers, including \tool{}, exhibit a more graceful decline in performance, maintaining much higher accuracy levels (72.79\% for HyperCam$^*$). Additionally, in terms of memory consumption, both {SVM} and {xgBoost} require significantly more memory than \tool{}. Even after being quantized to integer values, {SVM} could not fit in the MCU's flash memory.}

\reb{Neural network models, particularly those using 8-bit integer quantization, such as {MicroNets} and {MobileNetV3}, offer the highest accuracy levels (e.g., 98.69\% on MNIST for {MobileNetV3}) but at the cost of substantially higher memory usage and latency. For example, in terms of flash memory consumption, {MicroNets} requires over 500 KB of flash memory while {MobileNetV3} and {MCUNetV3} all use more than 1 MB of flash memory. On the other hand, HyperCam's most memory-efficient version ({count-sketch}) requires only 63 KB of flash memory and 22.25 KB of RAM. {MCUNetV3} exhibits the highest accuracy among all models, with near-perfect performance (e.g., 99.34\% on MNIST and 99.88\% on Face Detection). However, the trade-off comes in the form of substantially higher latency. The smallest {MCUNetV3} model has a latency of 6.7 seconds, while the larger version takes up to 46.71 seconds. In contrast, \tool{} maintains latencies under 0.3 seconds across all datasets.}

\begin{figure}[t!]
    \centering
    \includegraphics[width=\columnwidth]{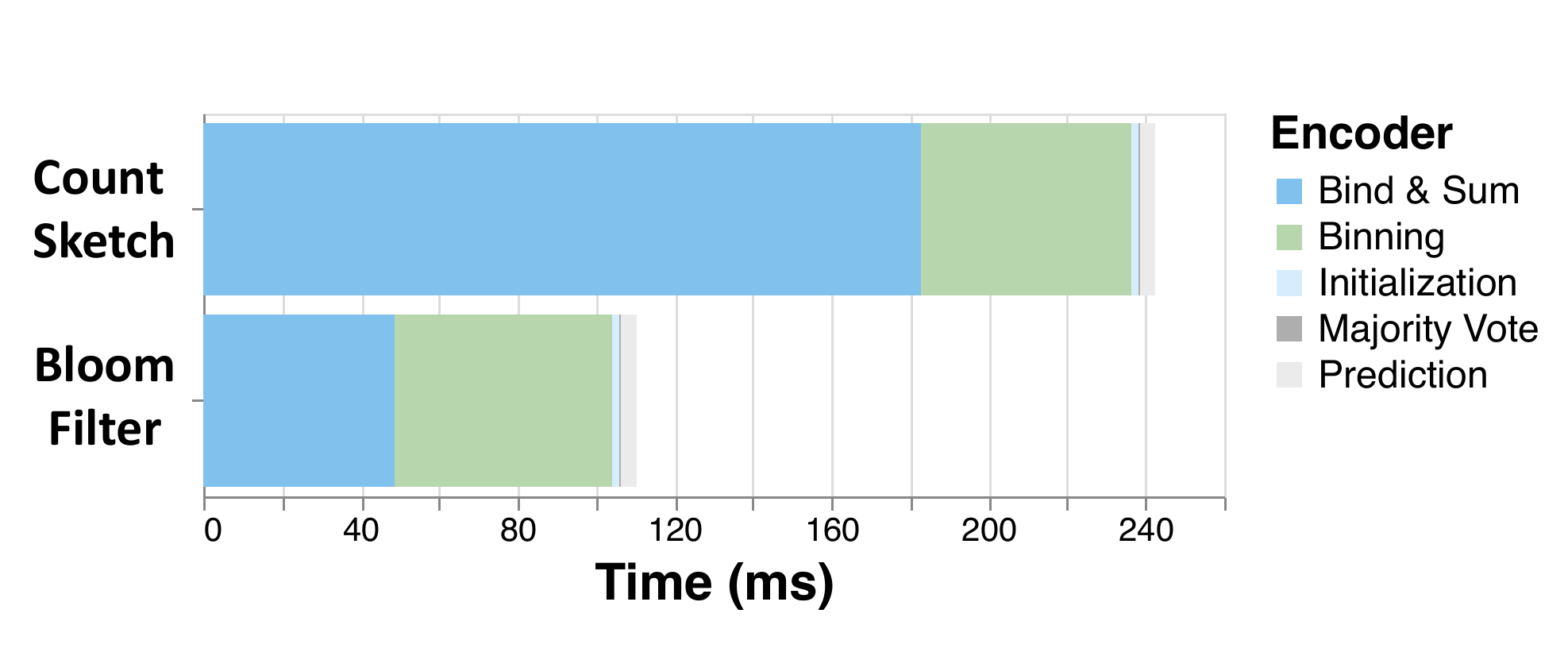}
    \caption{Latency Profiling of \tool{}.}
    \label{fig:latency}
    \vspace{-15pt}
\end{figure}



\prosehead{Memory Analysis.}
A breakdown of the flash memory is shown in Fig. \ref{fig:memory_breakdown}. For neural networks represented by MCUNetV3, memory is consumed by the TF Micro library and the model data (graph, weights, and parameters). On the other hand, in \tool{}, no library code is needed as it exclusively uses hardware-native operations such as addition, exclusive or, and comparison. Instead, \tool{}'s HD classifier only requires hypervectors stored as byte arrays. This aspect minimizes memory requirements and eases the process of adding new classifiers; each class addition consumes $n$ bits. In contrast, a new model in DNN comes with a model graph (about 100 KB) and weights that can be anywhere from hundreds of bytes to megabytes.

\prosehead{Latency Analysis.} Fig. ~\ref{fig:latency} shows the latency profiling of \tool when using Count Sketch and Bloom Filter. Observe that Bloom Filter outperforms Count Sketch for \texttt{bind \& sum}, improving overall latency. For the count-sketch method, the major processing time of \texttt{bind \& sum} is not binding and summing themselves but quantizing integer values to binary for binding. This process is omitted from the Bloom Filter backend, greatly reducing latency. Additionally, in both cases, bundling does not appear in Fig.~\ref{fig:latency} because it occurs in two stages: summation as part of \texttt{bind \& sum} and \texttt{majority vote}. That is, the result of binding is summed to the output element-wise and later evaluated for the majority.



\begin{table}[t!]
  \begin{center}
    \resizebox{\columnwidth}{!}{
    \begin{tabular}{|l|c|c|r|}
    \hline
      \textbf{Component} & \textbf{Active Current} & \textbf{Sleep Current} &
      \textbf{Voltage}\\
      \hline
      STM32U585I & 10.9 mA & 74.9 $\mu$A & 3.3V\\
      Himax HM01B0 & 2.5 mA & 1.3 mA & 2.8V\\
      nRF52840 Express & 7.2 mA & 1.4 mA & 3.8V\\
      \hline
    \end{tabular}
    }
    \caption{\tool Component Power Consumption.}
    \label{tab:power}
  \end{center}
  \vspace{-15pt}
\end{table}

\subsection{Power Analysis}
\begin{figure}[t!]
    \centering
    \includegraphics[width=\columnwidth]{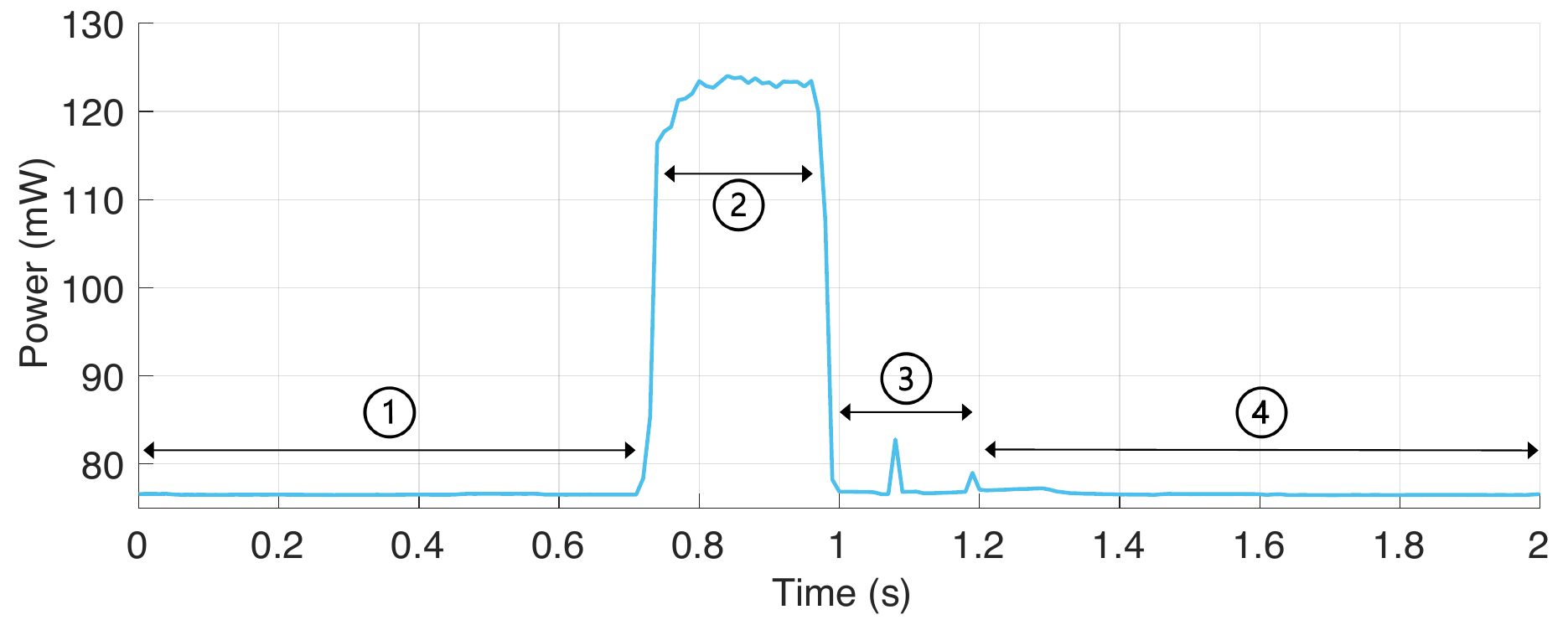}
    \caption{System Power Consumption. \normalfont{(1) camera plat- form in sleep mode (2) camera initialization, image capture, and inference, (3) data transmission, and (4) system returns to sleep mode.}}
    \label{fig:system-power}
    \vspace{-10pt}
\end{figure}

The power consumption of the wireless camera platform was evaluated using a Joulescope JS220 with 0.5 nA resolution, equivalent to 34 bits of dynamic range~\cite{joulescope}.  The average quiescent currents of the remaining components, the 3.3V STM32U585I and 2.8V Himax HM01B0 camera regulator are outlined in Table~\ref{tab:power}. Next, the system's total power consumption was evaluated during image capture, processing, and data transmission as shown in Fig.~\ref{fig:system-power}. The system was divided into key components during the power consumption measurement (MCU, camera, BLE module). The remaining B-U585I-IOT02A power can be derived by subtracting the power from the system's primary devices. When an event trigger occurs, the image sensor is activated, and the power consumption jumps to an average of 128 mW for image processing for 250 ms. Lastly, a data packet is transmitted to a base station over BLE for 200 ms. This equates to an average power consumption of 102 mW during active mode for 450 ms. Note that further power optimization could be done by
replacing the evaluation board's regulators with more efficient switching or LDO regulators.



\section{Related Work}

There has been recent work in energy-efficient cameras and machine learning to exploit the resource-constrained environments presented by IoT devices. Several paradigms govern the current space, including work to ease the load of transmitting image data and onboard computing.\\

\proseheadb{Onboard Computing.} There have been two lines of efforts to enable IoT devices onboard computing. The first is the \textit{TinyML} paradigm, where lightweight DNNs are developed for resource-constrained platforms. It uses frameworks such as TensorFlow Lite for Microcontrollers (TFLM) to quantize and prune weights to alleviate performance overhead, enabling various levels of hardware acceleration and model deployment \cite{manor2022custom, david2021tensorflow}. Works such as \cite{micronet, mcunet} use extensive network architecture search (NAS) to find an optimized neural network architecture for available memory resources. In \cite{lin2022device}, convolutional neural networks enable tiny on-device training with considerable memory limitations. Here, inference is performed, and over time, classifier weights are updated to improve performance with new input sensor data. These models provide concrete baselines for \tool{}'s performance, as compared in Section~\ref{sec:evaluation}.

\begin{table*}[t!]
  \small
  \begin{center}
    \resizebox{2\columnwidth}{!}{
    \scalebox{0.8}{
    \begin{tabular}{|l|c|c|c|c|c|}
    \hline
      \textbf{System} & \textbf{Active Power} & \textbf{Resolution} & \textbf{Frame Rate} & \textbf{Communication} & \textbf{Onboard Inference}\\ 
      \hline
      BackCam~\cite{josephson2019wireless} & 9.7 mW & 160×120 & 1 fps & backscatter & $\times$\\
      WISPCam ~\cite{naderiparizi2015wispcam} & 6 mW & 176×144 & 0.001 fps & backscatter & $\times$\\
      NeuriCam ~\cite{neuricam} & 85 mW & 640×480/740p & 15 fps & BLE & $\times$\\
      MCUNet ~\cite{lin2020mcunet} & N/A & 224×224 & N/A & N/A & $\checkmark$\\
      HyperCam & 128 mW & 160×120 &  8 fps & BLE & $\checkmark$\\
      \hline
    \end{tabular}
    }}
    \caption{IoT Camera Platforms. \normalfont{A comparison of \tool to existing camera platforms.}}
    \label{tab:camera}
  \end{center}
  \vspace{-15pt}
\end{table*}

On the other hand, there have been works exploring the use of hyperdimensional computing for onboard machine learning. However, hyperdimensional computing primarily focuses on time-series data, which does not have the same overhead as image processing~\cite{kleyko22}. An exception to this is \cite{hdface}, where HDC is used to enable emotion and face detection. This work uses histograms of gradients (HoGs) to extract the features of the image, which involves calculating the gradients of the pixels and binning them by angles. This front-end feature extractor not only introduces more computation load but also fails to resolve the inherent encoding complexity in HD image processing.



\proseheadb{Energy-efficient Wireless Cameras.} The advancements in low-power processors and image sensors have led to several recent works that focus on developing low-power wireless camera platforms for computer vision applications. In ~\cite{naderiparizi2015wispcam, naderiparizi2016wispcam}, a battery-free RFID camera is presented and evaluated for machine vision applications such as face detection. Here, subsampled images are transmitted to a base station that runs a face detection algorithm. If a face is detected, coordinates of windows within the image frame are transmitted back to WISPCam to retrieve higher-resolution images. In ~\cite{josephson2019wireless}, a low-power wireless camera platform is presented to support real-time vision applications where images are sent to a base station to perform image processing and face classification. Here, image compression is performed to help minimize overall latency and, in turn, reduce power consumption. More recently, ~\cite{neuricam} presents a deep-learning-based system for video capture from a low-power wireless camera platform. Similar to the aforementioned related work, the neural network processing runs at an edge server or in the cloud. Compared to prior work, \tool focuses on enabling onboard computer vision for energy-constrained wireless camera platforms rather than offloading image processing and classification to the edge or cloud. More closely related to \tool is ~\cite{mcunet}, which deploys a lightweight inference engine on an MCU system for onboard image processing. Table~\ref{tab:camera} compares \tool to similar IoT camera platforms.  Other work includes ~\cite{beetlecam,Afzal2022underwater}, which present ultra-low-power implementations of wireless camera platforms for extremely challenging environments such as underwater imaging and placing wireless cameras on insects. 

\section{Conclusion and Discussion}\label{sec:conclusion}


\reb{We introduce \tool{}, an onboard image classification pipeline that leverages hyperdimensional computing (HDC). To meet the stringent resource constraints of off-the-shelf MCUs, we propose original image encoding methods involving sparse binary vectors and on-the-fly codebook generations, which significantly reduce the number of operations and memory footprint. As a result, our system requires only about 60 KB of flash memory and 20 KB of RAM, achieving a latency of approximately 0.1 s while maintaining comparable accuracy across multiple classification tasks.}

\reb{A key advantage of \tool{} is its scalability and compatibility across a wide range of hardware platforms. Unlike most ML models that rely heavily on floating-point operations and require specialized hardware support, such as Neural Processing Units (NPUs) and Floating Point Units (FPUs), \tool{} only uses bits and bit operations. Moreover, \tool{} does not require any additional libraries, machine learning engines, Neural Architecture Search (NAS), or hardware-specific optimizations to reproduce results. That is, \tool{} can be easily ported to other families of MCUs.}

We highlight several future research directions:

\proseheadb{Onboard Training.} Data-driven models deployed in the real world must be able to reuse new data to refine themselves. This is crucial because the deployment data can vary greatly from the training data distributions, and unseen categories of data can appear. The system can adapt to these changes through onboard training. HDC offers an easy transition due to the ease of updating class hypervectors. New image hypervectors can be bundled into individual class hypervectors.

\proseheadb{Cloud Connection.} Currently, \tool{} connects to a gateway where the prediction result is transmitted and displayed. If the gateway establishes a connection to the cloud and pushes the data, remote users can access the data and monitor the results. The cloud can potentially store the bundling of thousands of image hypervectors as a model summary. This can be used to refine the model and provide future updates. 

\proseheadb{Multi-modal Sensors.} Modern IoT devices are equipped with a wide array of sensors. Providing multiple sensor data inputs to one intelligent model can lead to more accurate and quick decisions without the need to fuse this data with arbitrary algorithms. HD classifiers can easily fuse multiple types of sensor data because different data domains are all encoded to hypervectors. Examples of modalities include images and audio for speech recognition, EEG, and heart rate data for anomaly detection.

\proseheadb{Diverse Applications.} \tool{} can be applied to many different computer vision tasks. Many IoT camera systems target remote environments with limited power and connectivity, such as large-scale farms and underwater ocean profiling~\cite{Afzal2022underwater, farmbeats}. An energy-efficient onboard classifier can run in these environments and perform tasks such as pest detection, crop yield prediction, and wildlife monitoring. A large-scale deployment of \tool{} is yet to be tested.

\rev{\proseheadb{Advancements in HDC.} Current HD classifiers, including \tool{}, are well-suited for resource-constrained IoT applications but struggle with complex datasets, such as those with many classes or high-resolution images. \tool{} uses optimized encoding strategies, but future advances in HDC, such as better training algorithms and improved hypervector representations, can be integrated to tackle more complex tasks without losing the computational efficiency of \tool{}.}




\bibliographystyle{ACM-Reference-Format}

\end{document}